%% file: paper.tex
\pdfoutput=1
\documentclass[11pt]{article}

\usepackage[]{naacl2021}

\usepackage{times}
\usepackage{latexsym}

\usepackage[T1]{fontenc}
\usepackage[utf8]{inputenc}

\usepackage{microtype}

\usepackage{pgfplots}
\usepackage{tikz}
\usepackage{collcell}
\usepackage{booktabs}
\usepackage{wrapfig}
\usepackage{multicol}
\usepackage{comment}
\usepackage{multirow}
\usepackage{listings}
\usepackage{enumitem}
\usepackage{color, colortbl}
\definecolor{Gray}{gray}{0.9}

\usepackage[first=0,last=9]{lcg}

\usepackage{pgfplots}
\usepackage{tikz}
\usepackage[compat=1.0.0]{tikz-feynman}
\usepackage{tikz}
\usepackage{collcell}
\usepackage{caption} 
\usepackage{tikz}
\usepackage{amsmath}
\usetikzlibrary{fit,positioning,arrows}
\usetikzlibrary{shapes,arrows}
\usetikzlibrary{intersections}
\usepackage[T1]{fontenc}
\usepackage{caption}
\usepackage{booktabs}  % nice looking tables
\usepackage{siunitx}

\tikzset{
  main/.style={circle, minimum size = 5mm, thick, draw=black!80, node distance = 10mm},
  connect/.style={-latex, thick},
  box/.style={rectangle, draw=black!100}
}

\title{On the Use of Context for Predicting Citation Worthiness of Sentences in Scholarly Articles}
\author{Rakesh Gosangi, Ravneet Arora, Mohsen Gheisarieha, \\
  \textbf{Debanjan Mahata, Haimin Zhang} \\
  Bloomberg, New York \\
  \texttt{\{rgosangi,rarora62,mgheisarieha,dmahata\}@bloomberg.net,} \\
  \texttt{raymondzhang@gmail.com}\\}

\begin{document}
\maketitle
\begin{abstract}
\input{abstract.tex}
\end{abstract}

\section{Introduction and Background}
\label{introduction}
\input{introduction.tex}

\section{Methodology}
\label{method}
\input{method.tex}

\section{Experimental Work}
\label{experiments}
\input{experiments.tex}

\section{Conclusions}
\label{futurework}
\input{conclusions}

\bibliography{references}
\bibliographystyle{acl_natbib}

\end{document}

%% file: abstract.tex
% This is not the abstract you are looking for. 
In this paper, we study the importance of context in predicting the citation worthiness of sentences in scholarly articles. We formulate this problem as a sequence labeling task solved using a hierarchical BiLSTM model. We contribute a new benchmark dataset containing over two million sentences and their corresponding labels. We preserve the sentence order in this dataset and perform document-level train/test splits, which importantly allows incorporating contextual information in the modeling process. We evaluate the proposed approach on three benchmark datasets. Our results quantify the benefits of using context and contextual embeddings for citation worthiness. Lastly, through error analysis, we provide insights into cases where context plays an essential role in predicting citation worthiness. 

%% file: introduction.tex
Citation worthiness is an emerging research topic in the natural language processing (NLP) domain, where the goal is to determine if a sentence in a scientific article requires a citation\footnote{For example, in the first excerpt in Table \ref{tab:examples}, the goal is to predict that the first, third, and fourth sentences would require citations but the second does not.}. This research has potential applications in citation recommendation systems \cite{strohman2007recommending,kuccuktuncc2014diversifying,he2010context}, and is also useful for scientific publishers to regularize the citation process. Providing appropriate citations is critical to scientific writing because it helps readers understand how the current work relates to existing research.

Citation worthiness was first introduced by \cite{sugiyama2010identifying}, where the authors formulated as a sentence-level binary classification task solved using classical machine learning techniques like Support Vector Machines (SVMs). Subsequent works from \cite{farber2018cite,bonab2018citation} use similar approach but employ deep learning models like Convolutional Neural Networks (CNNs) and Recurrent Neural Networks (RNNs). More recently, Zeng et al. \cite{zeng2020modeling} proposed a Bidirectional Long short-term memory (BiLSTM) based architecture and demonstrated that context, specifically the two adjacent sentences, can help improve the prediction of citation worthiness. 
% \cite{zeng2020modeling} released another dataset based on articles from the PubMed Open Access Subset which not only contains sentences and labels but also the context: the preceding sentence, the next sentence, and their labels. 
% To our knowledge, these are the only published works that explicitly study the problem of sentence-level citation worthiness in scientific documents. 
% \cite{bonab2018citation} also released a dataset containing over one million sentences from research articles in the SEPID corpus \footnote{http://pars.ie/lr/sepid-corpus} and the corresponding sentence-level citation labels. 

Citation worthiness is closely related to citation recommendation (suggest a reference for a sentence in a scientific article), which is often approached as a ranking problem solved using models that combine textual, contextual, and document-level features \cite{strohman2007recommending,he2010context}. More recent works employ deep learning models \cite{huang2015neural, ebesu2017neural} and personalization \cite{cai2018three, yang2018lstm, cai2018generative}. Citation analysis \cite{athar2012context} and citation function \cite{teufel2006automatic,li2013towards,hernandez2017citation}, other closely related domains, aim to predict the sentiment and motivation of a citation respectively. Researchers have used many supervised approaches like sequence labeling \cite{abu2013purpose}, structure-based prediction \cite{athar2011sentiment}, and multi-task learning \cite{yousif2019multi} to address these problems.

\begin{table*}[t]
    \centering
    \small
    \begin{tabular}{|p{15cm}|}
        \hline
        \rowcolor{Gray}
        The baseline TTS systems in the project utilize the HTS toolkit which is built on top of the HTK framework \textbf{[Cite]}. The HMM-based TTS systems have been developed for Finnish, English, Mandarin and Japanese \textbf{[No Cite]}. The systems include an average voice model for each language trained over hundreds of speakers taken from standard ASR corpora, such as Speecon \textbf{[Cite]}. Using speaker adaptation transforms, thousands of new voices have been created and new voices can be added using a small number of either supervised or unsupervised speech samples \textbf{[Cite]}.\\
       % \hline
        %\rowcolor{Gray}        
        %We deal with this by defining a new two-stage integer programming formulation that identifies minimal grammars efficiently and effectively. CCGbank. \textbf{CCGbank was created by semiautomatically converting the Penn Treebank to CCG derivations} \textbf{[Cite]}. We use the standard splits of the data used in semi-supervised tagging experiments (e.g. ): sections 0-18 for training, 19-21 for development, and 22-24 for test \textbf{[Cite]}. \\
        \hline
        \rowcolor{Gray}
        % Researchers all over the world working on query based summarization are trying different directions to see which methods provide the best results.  
        The LexRank method addressed was very successful in generic multi-document summarization \textbf{[Cite]}. A topic-sensitive LexRank is proposed \textbf{[Cite]}. As in LexRank, the set of sentences in a document cluster is represented as a graph, where nodes are sentences and links between the nodes are induced by a similarity relation between the sentences \textbf{[No Cite]}.\\
        \hline
    \end{tabular}
    %\caption{These excerpts are obtained from \cite{kurimo2010personalising} and \cite{ravi2010minimized}. The actual citations have been removed but we the citation worthiness labels.}
    \caption{The excerpts are obtained from \cite{kurimo2010personalising} and \cite{chali2009complex}. The actual citations have been removed but we include citation worthiness labels.}
    \label{tab:examples}
\end{table*}

% \textbf{Move examples table here?}

In this paper, we want to investigate two research question about citation worthiness. First, we posit that citation worthiness is not purely a sentence-level classification task because the surrounding context could influence if a sentence requires a citation. This context could include not only adjacent sentences but also information about the section titles, paragraphs, and other included citations. Previous work \cite{bonab2018citation} has explored using the adjacent two sentences; we predict that citation worthiness models would improve with access to more contextual information. To pursue this hypothesis, we propose two new formulations: (a) sentence pair classification and (b) sentence sequence modeling. For the latter formulation, we propose a new hierarchical architecture, where the first layer provides sentence-level representations, and the second layer predicts the citation worthiness of the sentence sequence. We also introduce a new dataset mostly because the prior datasets \cite{bonab2018citation,zeng2020modeling} do not have sufficient contextual information to study this research question.

The second research objective is to understand if contextual embedding models would help citation worthiness. Recent developments in language modeling, specifically contextual embedding models \cite{liu2019roberta,devlin2019bert}, have already demonstrated significant improvements in various NLP research tasks. We expect to observe similar gains in citation worthiness. Following is a summary of the main contributions of this work:
\setlist{nolistsep}
\begin{itemize}[leftmargin=*, wide=0pt, noitemsep]
    \setlength\itemsep{0em}
    \item We propose two new formulations for citation worthiness: sentence-pair classification and sentence sequence modeling. 
    \item We contribute a new dataset containing significantly more context, and we expect it to serve as another benchmark. 
    \item Through rigorous experimental work, we demonstrate the benefits of sequential modeling and contextual embeddings for citation worthiness.
    \item We obtain new state-of-the-art results on three benchmark datasets.
\end{itemize}
% \setlength\itemsep{0em}

% To this end, we experiment with various contextual embedding models, and our results show that they obtain significant gains across all datasets.

%% file: method.tex
\subsection{Problem Statement}
Let $d = \{s_1, s_2, ..., s_n\}$ be a scientific article, where $s_i$ is the $i^{th}$ sentence. The problem of citation worthiness is to assign each sentence $s_i$ one of two possible labels $L = \{l_c, l_n\}$, where $l_c$ denotes that sentence requires a citation and $l_n$ means otherwise. We present three different formulations here to investigate our main research objectives.
% and compare our work with prior research. 

\begin{figure*}[t]
\centering
\includegraphics[width=14cm]{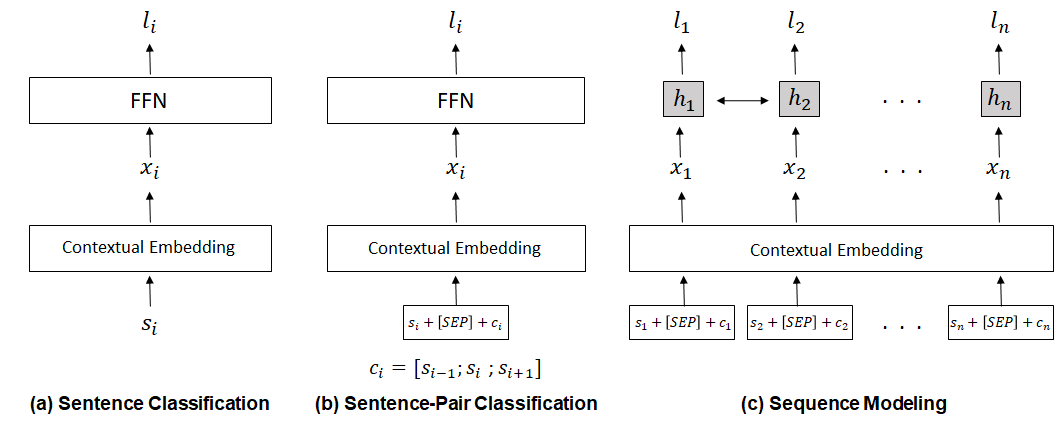}
\caption{The three proposed architectures.}
\label{fig:architectures}
\end{figure*}

\subsection{Sentence Classification}
% \label{sec:sentence}
Our first formulation (Figure \ref{fig:architectures} (a)) approaches citation worthiness as a sentence level classification task similar to prior works of \cite{bonab2018citation,farber2018cite}. Given a sentence $s_i$, we map it to a fixed-size dense vector $x_i$ using contextual embedding models (e.g. BERT \cite{devlin2019bert}). We then feed $x_i$ to a feed-forward layer to obtain the citation worthiness label. We fine-tune this entire architecture by optimizing the weights of the final layer.

\subsection{Sentence-Pair Classification}
\label{sec:sentence-pair}
Our second approach (Figure \ref{fig:architectures} (b)) is to formulate citation worthiness as a sentence-pair classification task, where the pair consists of the given sentence and a sentence-like representation of the context. Namely, for a given sentence $s_i$, we define the context $c_i$ as the concatenation of the previous $s_{i-1}$, $s_i$, and the next sentence $s_{i+1}$: 
\begin{equation}
\label{eq:1}
    c_i = [s_{i-1};s_i;s_{i+1}]
\end{equation}
We then concatenate $s_i$ with $c_i$ separated by the [SEP] token, pass it through the embedding layer to obtain a vector representation $x_i$. This vector is then passed through a feed-forward layer to obtain the class label. This approach is similar to \cite{zeng2020modeling}, where the authors used Glove embeddings \cite{pennington2014glove} to obtain sentence representations, and BiLSTMs for context representations. This formulation has also been used previously for question-answering \cite{devlin2019bert} and passage re-ranking \cite{nogueira2019passage}. In our sentence-pair classification approach, we defined $c_i$ to include only two adjacent sentences, but it could easily include more. However, if we included too many sentences, the context might be too long for most transformer-based models. 

% \label{sec:sentence-sequence}
% \textbf{TODO: Should we add a bit more motivation on why we are considering this model? Reviewers can say that we can use entire context in the sentence pair model. Possible motivation can be that Transformer based models have token length limitation. So it is not easy to include very long sequences of sentences. By modeling this as a sequence of Transformer models, the architecture becomes more scalable to applications where even longer context is needed while keeping the power of transformer models. Feel free to ignore it if you think this makes it more complicated.}
\subsection{Sentence Sequence Modeling}
The third formulation addresses citation worthiness as a sequence labeling task solved using a hierarchical BiLSTM architecture (Figure \ref{fig:architectures} (c)). We first map each sentence $s_i$ and context $c_i$ (eq. \ref{eq:1}) to a fixed-size dense vector $x_i$ using the same approach as in section \ref{sec:sentence-pair}. Thus the given document $d$ is represented as a sequence of vectors $x = \{x_1, x_2, ..., x_n\}$. We then feed these vectors to a BiLSTM model to capture the sequential relations between the sentences. The hidden state of the BiLSTM $h_i$ provides a vector representation for sentence $s_i$ that incorporates information from the surrounding sentences. Thus the sequence modeling approach captures long term dependencies between the sentences without us needing to explicitly encode them as extra features. We then use a feed-forward layer to map the BiLSTM output to the citation worthiness labels. We have also experimented with using only the sentence $s_i$ to construct the vector $x_i$. However, we observed that using the context $c_i$ helped improve the model performance.

%% file: experiments.tex
\begin{table*}[t]
    \centering
    \small
    \begin{tabular}{|l|r|r|r|r|}
        \hline
        \textbf{Model} & \textbf{P} & \textbf{R} & \textbf{F1} & \textbf{W-F1}  \\
        \hline
        \multicolumn{5}{|c|}{\textbf{SEPID-cite}} \\
        \hline
        CNN + Glove (f) &  0.196 & 0.269 & 0.227 & N/A \\
        CNN + w2v (b)  &  0.448 & 0.405 & 0.426 & N/A \\
        BiLSTM + Glove (z)  & \textbf{0.720} & 0.391  & 0.507 &  N/A \\
        %  & BERT & 0.51 & 0.66 & 0.579 &  \\
        %  & SciBERT & & & 0.6 &  \\
        SC + Roberta & 0.674 & \textbf{0.595} & \textbf{0.629} & 0.950  \\
        \hline
        \multicolumn{5}{|c|}{\textbf{PMOA-cite}} \\
        \hline
        BiLSTM + Glove (z) & 0.883 & 0.795 & 0.837 & N/A \\
        %  & BERT & 0.701 & 0.562 & 0.624 & 0.868 \\
        %  & SciBERT & 0.703 & 0.541 & 0.612 & 0.859 \\
        SC + Roberta & \textbf{0.887} & \textbf{0.814} & \textbf{0.849} & 0.942 \\
        \hline
        BiLSTM + Glove + context (z) & \textbf{0.907} & 0.811 & 0.856 & N/A\\
        %  & BERT & 0.715 & 0.608 & 0.657 & 0.872 \\
        %  & SciBert & 0.726 & 0.627 & 0.673 & 0.878 \\
        SPC + Roberta & 0.905 & \textbf{0.843} & \textbf{0.873} & 0.962 \\
        \hline
        \multicolumn{5}{|c|}{\textbf{ACL-cite}} \\
        \hline
        % SC & BERT & 0.8223 & 0.4686 & 0.597 & \\
        % & Scibert & 0.7981 & 0.5331 & 0.6392 &  \\
        SC + Roberta & 0.782 & 0.496 & 0.607  & 0.921\\
        % SPC & BERT & 0.8109 & 0.582 & 0.6776 & \\
        % & SciBERT  & 0.8038 & 0.5677 & 0.6654 & \\
        SPC + Roberta & 0.820 & 0.562 & 0.667 & 0.932\\
        % SSM & BiLSTM + BERT & 0.7907 & 0.6341 & 0.7038 & \\
        % & BiLSTM + SciBERT & 0.7996 & 0.6473 & 0.7155 &  \\
        SSM (4 sentences) + Roberta & \textbf{0.821} & 0.635 & 0.716 & 0.940\\
        SSM (8 sentences) + Roberta & 0.816 & 0.643 & 0.719 & 0.940\\
        SSM (16 sentences) + Roberta & 0.809 & 0.654 & 0.723 & 0.941\\
        SSM (16 sentences + section) + Roberta & 0.813 & \textbf{0.662} & \textbf{0.730} & \textbf{0.942}\\
        % SSM + Roberta & 0.809 & \textbf{0.654} & \textbf{0.723} & \textbf{0.941} \\
        \hline
    \end{tabular}
    \caption{SC: Sentence Classification, SPC: Sentence-Pair Classification, SSM: Sentence Sequence Modeling. (f) denotes that the numbers reported in \cite{farber2018cite}, (b) in \cite{bonab2018citation}, and (z) in \cite{zeng2020modeling}. P, R, and F1 are precision, recall, and F1 scores for the cite class ($l_c$). W-F1 is the weighted-F1 on entire test dataset.}
    \label{tab:results}
\end{table*}

% \textbf{Datasets}: 
\subsection{Datasets}
Prior works in citation worthiness presented two benchmark datasets: SEPID-cite \cite{bonab2018citation} and PMOA-cite \cite{zeng2020modeling}. SEPID-cite contains 1,228,053 sentences extracted from 10,921 articles\footnote{http://pars.ie/lr/sepid-corpus} but does not contain the source of the sentences (e.g. paper id) or the sentence order. PMOA-cite contains 1,008,042 sentences extracted from 6,754 papers from PubMed open access. PMOA-cite also contains the preceding sentence, the next sentence, and the section header. However, the authors of PMOA-cite did data splits at sentence-level, which means sentences from the same research paper could be part of train and test datasets. Since we cannot use either one of these datasets directly for sequence modeling, we chose to process the ACL Anthology Reference Corpus \cite{bird-etal-2008-acl}\footnote{https://www.aclweb.org/anthology/} (ACL-ARC) while preserving the sentence order, and then split the data at document-level.
% the authors of these datasets performed the train/test splits at the sentence-level

% We first considered using PMOA-cite or SEPID-cite to identify all sentences from a paper in a sequence, allowing us to incorporate more context into the models. However, we observed that the authors of these datasets performed the train/test splits at the sentence-level, which means sentences from the same research paper could be part of training and test datasets. Therefore, we chose to re-process the ACL-ARC corpus while preserving the sentence order, and then split the data at document-level. 
% : we cannot recover the sentence sequences without information leakage. 
% is significantly larger with more than 300 million sentences, but the authors from released a smaller version for benchmarking  

\begin{table}[h]
    \centering
    \small
    \begin{tabular}{|l|r|r|r|}
        \hline
          & \textbf{SEPID-cite} & \textbf{PMOA-cite} & \textbf{ACL-cite}  \\
        \hline  
        Articles & 10,921 & 6,754 & 17,440 \\
        Sections & - & 32,198 & 130,603 \\
        Paragraphs & - & 202,047 & 934,502 \\
        Sentences & 1,228,053 & 1,008,042 & 2,706,792 \\
        S w/o c  & 1,142,275 & 811,659 & 2,401,059 \\
        S w c & 85,778 & 196,383 & 305,733 \\
        A c/s & 131 & 132 & 141 \\
        A w/s & 22 & 20 & 22 \\
        \hline
    \end{tabular}
    \caption{Statistics of the three datasets. S w/o c: Sentences without citation, S w c: Sentences with citation, A c/s: Average characters per sentence, A w/s: Average words per sentene.}
    \label{tab:data_stats}
\end{table}

The latest version of ACL-ARC \cite{bird-etal-2008-acl}, released in 2015, contains 22,878 articles. Each article here contains the full text and metadata such as author names, section headers, and references. We first processed this corpus to exclude all articles without abstracts because they typically were conference cover sheets. Then, for each section in an article, we extracted paragraph information based on newlines. Then, we split the paragraphs into constituent sentences\footnote{https://github.com/fnl/segtok} and processed the sentences to obtain citation labels based on regular expressions (Appendix A). We then sanitized the sentences to remove all the citation patterns. 

The resulting new corpus (\textbf{ACL-cite}\footnote{Dataset available at: \textbf{https://zenodo.org/record/4651554}}) contained 2,706,792 sentences from 17,440 documents of which 305,733 sentences (11.3\%) had citations. Lastly, we performed document-level splits: training (10,464 docs, 1,625,268 sentences), validation (3,487 docs, 539,085 sentences), and test (3,487 docs, 542,081 sentences). To validate our citation regular expressions, we manually annotated a random sample of 500 sentences and observed only one error in the extracted labels. Table \ref{tab:data_stats} provides some basic statistics on the three datasets. 

\begin{table*}[ht]
    \centering
    \small
    % \sisetup{round-mode=places}
    % \begin{tabular}{|l|S[round-precision=3]|S[round-precision=3]|S[round-precision=3]|S[round-precision=3]|S[round-precision=3]|S[round-precision=3]|}
    \begin{tabular}{|l|r|r|r|r|r|r|}
        \hline
        & \multicolumn{3}{|c|}{\textbf{SC}} & \multicolumn{3}{|c|}{\textbf{SSM}} \\
        \hline
        Section & P & R & F1 & P & R & F1 \\
        \hline
        Abstract & 0.340 & 0.505 & 0.407 & 0.543 & 0.576 & \textbf{0.559} \\
        Acknowledgments & 0.874 & 0.361 & 0.511 & 0.759 & 0.480 & \textbf{0.588} \\
        Conclusion & 0.585 & 0.459 & 0.515 & 0.711 & 0.560 & \textbf{0.626} \\
        Evaluation & 0.770 & 0.538 & 0.633 & 0.808 & 0.659 & \textbf{0.726} \\
        Introduction & 0.833 & 0.514 & 0.636 & 0.831 & 0.645 & \textbf{0.726} \\
        Methods & 0.791 & 0.525 & 0.631 & 0.803 & 0.650 & \textbf{0.718} \\
        Related Work & 0.901 & 0.707 & 0.792 & 0.918 & 0.827 & \textbf{0.870} \\
    \hline
    \end{tabular}
    \caption{Comparison of the F1 score of SC and SSM models by section.}
    \label{tab:section-analysis}
\end{table*}

\subsection{Experimental settings}

We applied the sentence classification (SC) model on all three datasets, sentence-pair classification (SPC) model on PMOA-cite and ACL-cite, and sentence sequence modeling (SSM) approach on ACL-cite. This is because SEPID-cite does not have any context to apply SPC or SSM, and PMOA-cite does not have sufficient context for SSM. 

To obtain sentence representations, we also explored the idea of pooling word-level embeddings obtained using CNNs. However, we observed no significant difference in the model performance when compared to using the [CLS] token. We also experimented with the choice of contextual embeddings: BERT \cite{devlin2019bert}, SciBERT \cite{beltagy2019scibert}, Roberta \cite{liu2019roberta}, and XLnet \cite{yang2019xlnet} and observed that the Roberta model consistently gave the best results; therefore, we only report those numbers.
% and decided to use [CLS] token since it simplifies the architecture. 
% We explored using the [CLS] token or pooling the word-level embeddings using CNNs to obtain sentence representations from the contextual embedding models. We observed no significant difference in the model performance and decided to use [CLS] token since it simplifies the architecture.  
% We also empirically verified that $c_i$ was beneficial to the SSM models. 

We used a batched training approach for the SSM models: split each article into sequences of $m$ with an overlap of $m/2$ sentences. For example, consider a document with 32 sentences and $m=16$, we create three training sequences; first sequence: sentences 1 to 16, second sequence: sentences 9 to 24, and so on. During inference, for a given sentence, we include the preceding $m/2$ sentences and the succeeding $m/2-1$ sentences\footnote{We used zero-padding in cases with insufficient context, e.g., beginning or end of a document.}. We trained and evaluated models at different values of $m={4,8,16}$. We trained all the models using the Adam optimizer with a batch size of 16, a learning rate of 1e-5, a maximum of 4 epochs to optimize for cross-entropy loss. The hidden layers in the BiLSTM models were set to 128 units. The models were trained on a GPU machine with 6 cores and each training epoch took approximately 4 hours. More details on the experimental settings are available in the Appendix. 

% Since all the datasets are imbalanced, we downsampled the negative class ($l_n$) to 20\%.
% As shown in Table \ref{tab:data_stats}, the citation-worthiness datasets are imbalanced; therefore we downsampled the negative class ($l_n$) for training. 

% Training parameters: batch size, epochs, annealing factor, patience, learning rate, BiLSTM units, dropout
% Validation dataset was used for hyper-parameter tuning.

\subsection{Results}
Table \ref{tab:results} summarizes the results in terms of the precision, recall, F1 score for $l_c$, and overall weighted F1 score. The baseline numbers reported here are either from prior works \cite{farber2018cite,bonab2018citation,zeng2020modeling} or based on architectures very similar to those used in these prior works. On the SEPID-cite dataset, our SC model obtained significantly better performance than the state-of-the-art results from \cite{zeng2020modeling} with the F1 score increasing by more than 12\%. On the PMOA-cite dataset, we obtain an F1 gain of 1.2\% for sentence-level and 1.7\% for contextual models. We indicate that the numbers from \cite{zeng2020modeling} use additional hand-crafted contextual features, including labels of surrounding sentences, but our models only use textual features.
% We indicate here that the numbers reported from  also use additional hand-crafted contextual features (including labels of surrounding sentences) but our models only use textual features. 

The results on the ACL-cite dataset clearly show the importance of context in this domain. The use of surrounding two sentences boosted the performance by nearly 6\% points, and the performance continues to improve with added context increasing by another 5.6\% points for 16 sentences. The model performance improves by another 0.7\% with the inclusion of section headers in the context. 
Table \ref{tab:section-analysis} compares the performance of the SC and SSM models for different sections in the papers. The F1 score improves for all but most prominent for \textit{Abstract} and \textit{Conclusion} sections because of significant improvements in the precision. 
% where citations are usually uncommon. We also observe significant gains in precision for these two sections. 
% \begin{figure}
% \includegraphics[width=\linewidth]{section_analysis.jpg}
% \caption{Section-wise improvements gained by introducing context}
% \label{fig:section-analysis}
% \end{figure}
% To better understand the source of improvements achieved by the introduction of context, we have plotted section-wise improvements gained by the Sequence Modeling architecture over the no-context Sentence Classification model. We observe an improvement in precision for categories such as Abstract and Conclusion where the citations are not very common. The context provides section level information to the model and prevents it from making wrong predictions. There is also a uniform gain in recall and F-1 scores across all the section categories. This highlights the general effectiveness of the proposed model for different sections of the paper.  

    % \begin{tabular}{
    %   S[round-precision=2]
    %   S[round-precision=3]
    % }\toprule
    %   1.98185942 & 0.14495331 \\ \bottomrule
    % \end{tabular}

\pgfplotstableread[row sep=\\,col sep=&]{
    section & sc & ssm \\
Abstract & 0.406652361 & 0.558861578 \\
Acknowledgments & 0.510526316 & 0.587699317 \\
Conclusion & 0.514775161 & 0.626176219 \\
Evaluation & 0.63333859 & 0.726172912 \\
Introduction & 0.635664824 & 0.726015907 \\
Methods & 0.631259051 & 0.718175681 \\
Related Work & 0.792373988 & 0.870172858 \\
}\sectionfdist

% \pgfplotsset{width=8cm, height=4cm}
% \begin{figure}
% \centering
% \small
% \begin{tikzpicture}
% \begin{axis}[
%     ybar,
%     bar width=3pt,
%     symbolic x coords={Abstract, Acknowledgments, Conclusion, Evaluation, Introduction,  Methods, Related Work},
%     xtick=data,
%     ylabel={f1},
%     ymin=0,ymax=1.0,
%     x tick label style={rotate=30,anchor=east},
%     legend style={at={(0.3,0.9)}, anchor=north,legend columns=0, font=\tiny}
% ]
%     \addplot table[x=section,y=sc]{\sectionfdist};
%     \addplot table[x=section,y=ssm]{\sectionfdist};
% \legend{SC, SSM}
% \end{axis}
% \end{tikzpicture}
% \caption{Comparison of the f1 scores of SC and SSM models by section.}
% \label{fig:section-analysis}
% \end{figure}

\pgfplotstableread[row sep=\\,col sep=&]{
    section & precision & recall & F1 \\
Abstract & 0.202498128 & 0.070666667 & 0.152209218 \\
Acknowledgments & -0.115050344 & 0.118959108 & 0.077173001 \\
Conclusion & 0.125480001 & 0.100152905 & 0.111401058 \\
Evaluation & 0.03827153 & 0.121350121 & 0.092834322 \\
Introduction & -0.002069082 & 0.130720633 & 0.090351083 \\
Methods& 0.011493444 & 0.124618308 & 0.08691663 \\
Related Work & 0.016934356 & 0.12004662 & 0.07779887 \\
}\sectiondist

\subsection{Subjective Analysis}
We observed some interesting trends during the error-analysis of the SC and SSM models. We categorized these trends into three groups and selected an example from each to illustrate the impact of context (Table \ref{tab:examples}). 

\setlist{nolistsep}
\begin{itemize}[leftmargin=*, wide = 0pt, noitemsep]
\setlength\itemsep{0em}
\item \textit{Prior works}: In the first excerpt, the last sentence could be interpreted as the author's contribution if no context was available. The preceding sentences in the paragraph seem to help the model understand that this sentence requires a citation because it refers to prior work. 

% \item \textbf{Identify the subsection under which the statement has been made}.
%\item Section headers: In the second excerpt, the second sentence (\textit{CCGBank was created...} ) could easily be interpreted as a concluding sentence. The context around the sentence provides information for the the model to infer the section correctly and therefore the correct label.

\item \textit{Sections}: In the second excerpt, the second sentence could be interpreted as an introduction or conclusion. Once again, the context provides information to infer the section correctly and, therefore, the correct label.

\item \textit{Topic sentences}: Context is essential to understand if a sentence is the first statement about a topic, typically when researcher provide citations, or continuation of a discussion. In the second excerpt, the model does not predict $l_c$ for the last sentence because the authors already introduced the concept \textit{LexRank} in previous sentences.
\end{itemize}

%% file: conclusions.tex
In this paper, we study the impact of context and contextual models on citation worthiness. We propose two new formulations for this problem: sentence-pair classification and sentence sequence modeling. We contribute a new benchmark dataset with document-level train/dev/test splits, which enables to incorporate contextual information better. We propose a hierarchical BiLSTM approach for sequence modeling, but we could also consider a transformer-based approach and further improve with a CRF layer. Likewise, we also want to consider some of the newer language models \cite{zaheer2020big,beltagy2020longformer} that handle longer sentences. 

We expect citation worthiness would be an important part of developing writing assistants for scientific documents. We studied the citation worthiness of sentences in scholarly articles in this paper, but we believe these findings are relevant to other domains like news, Wikipedia, and legal documents.